%% file: main.tex
\documentclass[letterpaper,conference]{IEEEtran}
\usepackage[numbers,sort&compress]{natbib}
\usepackage{graphicx}
\usepackage{dirtytalk}
\usepackage{etoolbox}
\usepackage{booktabs}
\usepackage{multirow}

\usepackage[mathscr]{euscript}

\usepackage{caption}
\usepackage{subcaption}

\usepackage[table,xcdraw]{xcolor}
\makeatletter

\patchcmd{\@makecaption}
 {\\}
 {.\ }
 {}
 {}
 
\ifCLASSINFOpdf
\else
\fi
\usepackage{amsmath}
\usepackage{amssymb}

\usepackage{algorithm}
\usepackage{algpseudocode}
\usepackage{enumerate}
\usepackage[shortlabels]{enumitem}

\usepackage{threeparttable}
\usepackage{scalerel}
\usepackage{tikz}
\usetikzlibrary{svg.path}

\definecolor{orcidlogocol}{HTML}{A6CE39}
\tikzset{
    orcidlogo/.pic={
        \fill[orcidlogocol] svg{M256,128c0,70.7-57.3,128-128,128C57.3,256,0,198.7,0,128C0,57.3,57.3,0,128,0C198.7,0,256,57.3,256,128z};
        \fill[white] svg{M86.3,186.2H70.9V79.1h15.4v48.4V186.2z}
        svg{M108.9,79.1h41.6c39.6,0,57,28.3,57,53.6c0,27.5-21.5,53.6-56.8,53.6h-41.8V79.1z M124.3,172.4h24.5c34.9,0,42.9-26.5,42.9-39.7c0-21.5-13.7-39.7-43.7-39.7h-23.7V172.4z}
        svg{M88.7,56.8c0,5.5-4.5,10.1-10.1,10.1c-5.6,0-10.1-4.6-10.1-10.1c0-5.6,4.5-10.1,10.1-10.1C84.2,46.7,88.7,51.3,88.7,56.8z};
    }
}

\newcommand\orcidicon[1]{\href{https://orcid.org/#1}{\mbox{\scalerel*{
                \begin{tikzpicture}[yscale=-1,transform shape]
                \pic{orcidlogo};
                \end{tikzpicture}
            }{|}}}}

\usepackage{hyperref}

\begin{document}

\title{High-Quality Medical Image Generation from Free-hand Sketch}

\author{\IEEEauthorblockN{Quan Huu Cap, Atsushi Fukuda}

% \IEEEauthorblockA{quan.cap@aillis.jp\quad atsushi.fukuda@aillis.jp\quad iyatomi@hosei.ac.jp}
\IEEEauthorblockA{AI Development Department, Aillis, Inc., Tokyo, Japan\\}
\IEEEauthorblockA{Email: \{quan.cap, atsushi.fukuda\}@aillis.jp}
}

\maketitle
% \thispagestyle{plain}
% \pagestyle{plain}
% ABSTRACT
\begin{abstract}
    \input{00_Abstract}
\end{abstract}

% KEYWORDS

% For peer-review papers, this IEEEtran command inserts a page break and
% INTRODUCTION
\section{Introduction}
    \input{01_Introduction}

\section{Related work}
    \input{02_Related_work}
    
\section{Proposed method}
    \input{03_Proposed_method}

\section{Experimental settings}
    \input{04_Experiment_settings}

\section{Results and discussion}
    \input{05_Result_Discussion}

\section{Conclusion}
    \input{06_Conclusion}
    
\section*{Acknowledgment}
We would like to express our sincere thanks to all members at Aillis Inc. who contributed to this project, especially Hiroshi Yoshihara (M.Sc.) and Wataru Takahashi (M.Eng.) for their valuable comments and feedback. 

\nocite{*}
\footnotesize{
\bibliographystyle{IEEEtran}
\bibliography{reference}
}

\end{document}

%% file: 00_Abstract.tex
Generating medical images from human-drawn free-hand sketches holds promise for various important medical imaging applications. 
Due to the extreme difficulty in collecting free-hand sketch data in the medical domain, most deep learning-based methods have been proposed to generate medical images from the synthesized sketches (e.g., edge maps or contours of segmentation masks from real images). 
However, these models often fail to generalize on the free-hand sketches, leading to unsatisfactory results. 
In this paper, we propose a practical free-hand sketch-to-image generation model called Sketch2MedI that learns to represent sketches in StyleGAN's latent space and generate medical images from it. 
Thanks to the ability to encode sketches into this meaningful representation space, Sketch2MedI only requires synthesized sketches for training, enabling a cost-effective learning process. 
Our Sketch2MedI demonstrates a robust generalization to free-hand sketches, resulting in high-quality and realistic medical image generations. 
Comparative evaluations of Sketch2MedI against the pix2pix, CycleGAN, UNIT, and U-GAT-IT models show superior performance in generating pharyngeal images, both quantitative and qualitative across various metrics. \\

%% file: 01_Introduction.tex
Human-drawn free-hand sketches are an effective way of visualizing target objects or concepts. 
In the domain of medical imaging, the synthesis of medical images from sketches offers potential for various important applications. 
One such application is for patient communication, where physicians can interactively explain conditions or medical concepts to patients through the generated images from free-hand sketches. 
Furthermore, sketch-to-image generation enables the application of sketch-based medical image retrieval. 
This approach allows medical professionals to sketch and retrieve images based on their partial memory, enhancing the efficiency of image retrieval processes. 

In recent years, many deep learning-based methods have been proposed to generate images from free-hand sketches and achieved remarkable results \cite{chen2018sketchygan, gao2020sketchycoco, ham2022cogs, koley2023picture}. 
However, unlike general computer vision tasks, collecting medical images and the corresponding sketches to build such models is very difficult and time-consuming since it requires a lot of expert knowledge as well as drawing effort. 
To mitigate this issue, some studies proposed using synthesized sketches such as edge maps \cite{isola2017image, li2019linestofacephoto, chen2020deepfacedrawing} or semantic contours (i.e., boundaries of ground-truth semantic masks) \cite{li2020deepfacepencil} from the target photos as pseudo sketch-replacement for model training. 
However, human-drawn free-hand sketches of objects are usually in different levels of abstractions and deformations, whereas synthesized sketches ideally align with target image boundaries. 
Because of this large domain difference, images generated from free-hand sketches by those models easily follow the abstract boundaries, yielding unrealistic results. 
Moreover, some methods are data-dependent and thus, cannot be applied to other modalities such as medical data \cite{chen2020deepfacedrawing, chen2021deepfaceediting}. 
We hypothesize that if there is a way to represent the synthesized sketches in a more abstract manner, the generated results from free-hand sketches could be significantly improved. 

It has been well-known that the latent space in StyleGAN \cite{karras2019style} expresses semantically rich representations, allowing different levels of abstractions to be manipulated. 
In recent years, many studies have proposed encoding real images into this latent space to facilitate diverse and expressive image manipulations \cite{abdal2019image2stylegan, tov2021designing, alaluf2022hyperstyle, liu2023delving}. 
Inspired by these works, in this paper, we propose an encoder-decoder-based model called Sketch2MedI to encode sketches into StyleGAN’s meaningful latent space and subsequently, generate medical images from this manifold to ensure photorealistic quality. 
To enable cost-effective learning, we propose using a sketch synthesizer which is a trained segmentation model to synthesize sketch data (i.e., predicting semantic contours) from real medical images to train our Sketch2MedI model. 
Despite training entirely on the synthesized sketch data, we demonstrate that our proposal generalizes very well in generating pharyngeal images from human-drawn free-hand sketches. 
In comparison to the pix2pix \cite{isola2017image}, CycleGAN \cite{zhu2017unpaired}, UNIT \cite{liu2017unsupervised}, and U-GAT-IT \cite{kim2020u} models, the generated results by our Sketch2MedI showed substantial improvements, both quantitatively and qualitatively. 
% Figure 1
\input{figures/tex_files/Fig_1}

%% file: figures/tex_files/Fig_1.tex
%Figure 1
\begin{figure*}[!t]
\centering
\includegraphics[width=0.9\textwidth]{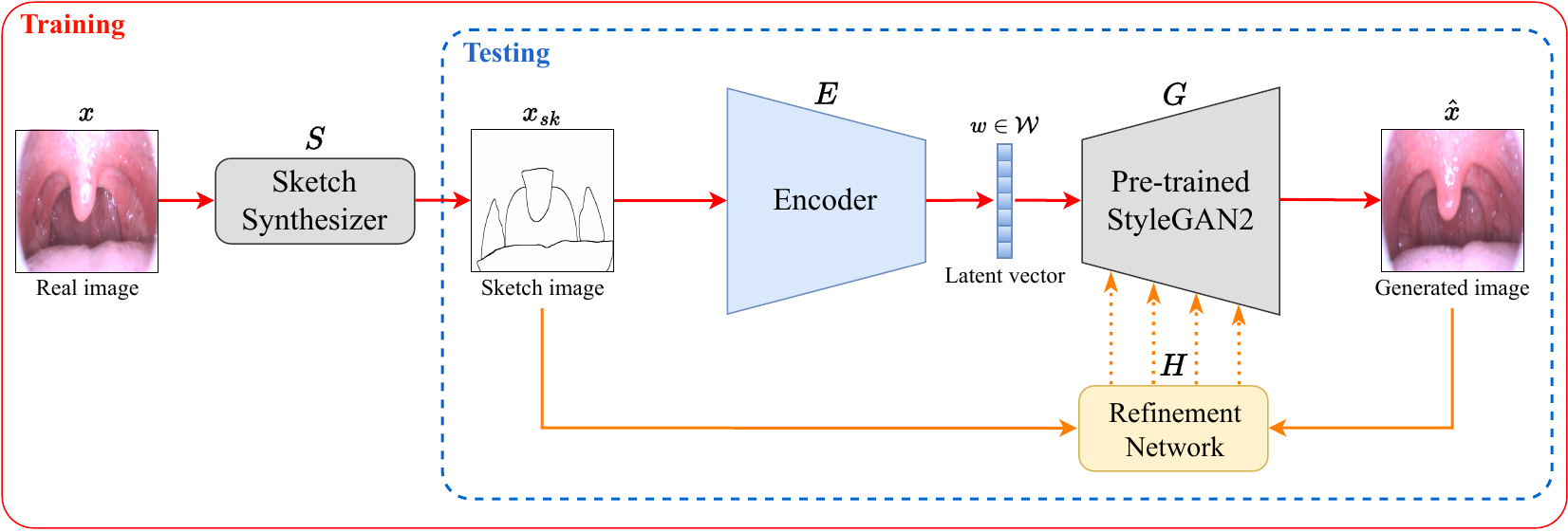}
\caption{
    The overview of our proposed Sketch2MedI. 
    At test time, we only input the human-drawn free-hand sketches into the model (ignore using $S$). 
}
\label{fig:fig_1}
\end{figure*}

%% file: 02_Related_work.tex
In medical imaging, several studies have been proposed utilizing sketch to image generation \cite{zhang2019skrgan,liang2022sketch,toda2022lung}. 
Zhang \textit{et al.} \cite{zhang2019skrgan} proposed generating fake edge maps from noises and using them to generate various medical image modalities. However, using edge maps differs largely from free-hand sketches. 
Moreover, this system lacks user control as the generated results are from random noises. 
Liang \textit{et al.} \cite{liang2022sketch} designed a sketch-guided model for realistic and editable ultrasound image synthesis. 
Nevertheless, edge maps from real images must be extracted beforehand to synthesize the results. 
Furthermore, training their model requires segmentation annotations for all images, resulting in substantial costs. 
% In contrast, our Sketch2MedI only requires free-hand sketches to generate the results at test time. 
In contrast, we train our Sketch2MedI using sketch data synthesized through a segmentation model that was trained on a subset of the dataset, providing a more cost-effective option. 
Toda \textit{et al.} \cite{toda2022lung} trained a model to generate lung cancer computed tomography (CT) images from edge maps for data augmentation. 
Despite showing promising results on free-hand sketches, the practicality of their model remains uncertain since at test time, each sketch needs to be carefully adjusted with visual feedback from physicians to avoid missing relevant information in the drawing, thus introducing complexity to the process. 
Moreover, their model was evaluated on a notably limited dataset with only 20 images. 
Therefore, the model's generalizability on diverse styles of free-hand sketches remains unknown. 

%% file: 03_Proposed_method.tex
\subsection{Overview}\label{sec:sec_2_1}
Our work is built on the idea of StyleGAN inversion \cite{abdal2019image2stylegan} where real images are encoded into the expressive and meaningful latent space $\mathcal{W}$ for image manipulation. 
In StyleGAN, the latent vector $w \in \mathcal{W}$ is mapped from a random noise vector $z \in Z=\mathcal{N}(\mu,\sigma^2)$ via a non-linear mapping function. The main objective of this study is to encode sketch images into this $\mathcal{W}$ space to enable realistic medical image generation. 

We designed our Sketch2MedI model as an encoder-decoder network to generate pharyngeal images from human-drawn free-hand sketches. 
Fig. \ref{fig:fig_1} illustrates the overview of our proposal. 
Given the real pharyngeal image $x$, the sketch synthesizer $S$, and the pre-trained StyleGAN2 \cite{karras2020analyzing} generator $G$, we first apply $S$ to synthesize the grayscale sketch image $x_{sk}=S(x)$. 
Then, the encoder $E$ encodes $x_{sk}$ into the $\mathcal{W}$ space to obtain the latent vector $w=E(x_{sk})\in \mathcal{W}$. 
From this, the generator $G$ is used to generate the output image $\hat{x}=G(w)$. 

In general, using the encoder $E$ alone (i.e., without $H$ employed) is sufficient to generate realistic images. 
However, those generated results occasionally are not well-aligned with pharyngeal image object boundaries. 
To further improve the generated quality, the refinement network $H$ takes the $x_{sk}$ and $\hat{x}$ images as inputs, and then gradually adjusts the weights of the generator $G$ in an \textit{iterative} manner, refining the quality of the output $\hat{x}$. 

We train our Sketch2MedI model in two phases. 
In the first phase, the encoder $E$ is trained without employing the refinement network $H$. 
In the second phase, the refinement network $H$ is trained with the encoder $E$ frozen. 
Note that the $S$ and $G$ networks remain frozen during both phases. 
At test time (Fig. \ref{fig:fig_1} (dashed blue box)), we instead use the human-drawn free-hand sketches as inputs to our Sketch2MedI model to generate pharyngeal images (ignore using $S$). 

\subsection{The sketch synthesizer}
Since collecting medical images and their corresponding free-hand sketches is a challenging task, we propose the sketch synthesizer $S$, which is a trained instance segmentation model to instead synthesize sketch data (i.e., semantic contours) from real images for training the Sketch2MedI. 
We chose to utilize semantic contours since they are less noisy than edge maps and have a closer appearance to free-hand sketches. 

Specifically, $S$ predicts the segmentation masks from a real pharyngeal image $x$, the synthesized grayscale sketch $x_{sk}$ is then formed by extracting the object contours from all predicted instance masks. 
Our sketch synthesizer $S$ is based on the SOLOv2 model \cite{wang2020solov2} and was trained on a subset of the dataset. 
Compared to collecting medical free-hand sketches, using $S$ to synthesize training data enables much more cost-effective learning. 

\subsection{The encoder}
The encoder $E$ is designed to encode the sketch $x_{sk}$ into StyleGAN’s latent space $\mathcal{W}$ to enable realistic medical image generation. 
We adopt the ResNet50-based \cite{he2016deep} encoder from \cite{tov2021designing} which takes $x_{sk}$ as input and outputs a latent vector $w \in \mathbb{R}^{512}$. 
A similarity loss is employed to train $E$, which ensures low distortion between real image $x$ and generated image $\hat{x}$. 
Following \cite{tov2021designing}, the similarity loss $\mathcal{L}_\mathrm{sim}$ includes a cosine similarity loss $\mathcal{L}_\mathrm{cos}$, a $L_2$ loss, and an LPIPS perceptual loss $\mathcal{L}_\mathrm{LPIPS}$ \cite{zhang2018unreasonable}. 
The $\mathcal{L}_\mathrm{sim}$ is as follows: 
%% Equation (1)
\begin{multline}
\label{eq:1}
\mathcal{L}_\mathrm{sim}=\lambda_{l2}L_2(x,\hat{x})+\lambda_{lpips}\mathcal{L}_\mathrm{LPIPS}(x,\hat{x})+\lambda_{cos}\mathcal{L}_\mathrm{cos}(x,\hat{x}),
\end{multline}
where $\lambda_{l2}, \lambda_{lpips}, \lambda_{cos}$ are the loss coefﬁcients. 
The objective of $\mathcal{L}_\mathrm{cos}$ is to minimize the cosine similarity between the feature embeddings of $x$ and $\hat{x}$. 
The $\mathcal{L}_\mathrm{cos}$ is defined as: 
%% Equation (2)
\begin{equation}
\label{eq:2}
\mathcal{L}_\mathrm{cos}=1-\mathrm{cos}(C(x),C(\hat{x})),
\end{equation}
where $C$ is a pre-trained self-supervised MOCOv2 model \cite{chen2020improved}. 

In addition, the encoder $E$ is trained with an adversarial loss \cite{goodfellow2014generative} that encourages the encoded latent vector $w$ to lie close to the latent space $\mathcal{W}$ of StyleGAN \cite{tov2021designing}. 
A 4-layer fully connected network from \cite{nitzan2020face} is applied as the latent discriminator $D$ to discriminate between real samples of the $\mathcal{W}$ space (generated from noise vector $z \in Z$) and the encoder’s outputs. 
The adversarial loss is the GAN loss \cite{goodfellow2014generative} with $\mathrm{R}_1$ regularization \cite{mescheder2018training}: 
%% Equation (3)
\begin{multline}
\label{eq:3}
\mathcal{L}_\mathrm{adv}^D=-\mathbb{E}_{w\sim \mathcal{W}}[\log D(w)]-\mathbb{E}_{x\sim p(x)}[\log (1-D(E(x)))]+\\ \frac{\gamma}{2}\mathbb{E}_{w\sim \mathcal{W}}[\left\|\nabla_w D(w)\right\|_2^2],
\end{multline}
%% Equation (4)
\begin{equation}
\label{eq:4}
\mathcal{L}_\mathrm{adv}^E=-\mathbb{E}_{x\sim p(x)}[\log D(E(x))].
\end{equation}
Finally, the total objective loss $\mathcal{L}_E$ for training the encoder $E$ is: 
%% Equation (5)
\begin{equation}
\label{eq:5}
\mathcal{L}_E=\mathcal{L}_\mathrm{sim}+\lambda_{adv}\mathcal{L}_\mathrm{adv}^E.
\end{equation}
Here, $\gamma$ and $\lambda_{adv}$ are the loss coefficients. 
For more details on the architecture and implementation of the encoder $E$, please refer to the literature \cite{tov2021designing}. 

\subsection{The refinement network}
We build our refinement network $H$ based on the hypernetwork from \cite{alaluf2022hyperstyle}. 
The task of $H$ is to further improve the quality of the generated image $\hat{x}$ in an \textit{iterative} manner. 
The network $H$ receives a 4-channel image (concatenated $x_{sk}$ and $\hat{x}$) and predicts a set of weight offsets with respect to the weights of the pre-trained StyleGAN2 \cite{karras2020analyzing} generator $G$. 
We perform several steps through $H$ for a single sketch $x_{sk}$ (e.g., $T$ steps). 
Each refinement step allows $H$ to gradually reﬁne its predicted weight offsets, helping the generator $G$ to produce a more precise result. 

% Let the weights of a convolution layer $l$ in $G$ be $\theta^{\lbrack l\rbrack}\in\mathbb{R}^{k_l\times k_l\times d_l^{in}\times d_l^{out}}$ where $k_l \times k_l$ is the kernel size; $d_l^{in},d_l^{out}$ are input and output channels, respectively. 
% The set of weights to be refined in $G$ is denoted as ${\{\theta\}}_{l=1}^M$, with $M$ is the number of layers. 
% The network $H$ is based on the ResNet34 model \cite{he2016deep} along with $M$ refinement blocks \cite{alaluf2022hyperstyle}. 
% For a convolution layer $l$, the corresponding refinement block outputs an offset $\Delta^{\lbrack l\rbrack}\in\mathbb{R}^{1\times 1\times d_l^{in}\times d_l^{out}}$. 
% The offset $\Delta^{\lbrack l\rbrack}$ is then replicated to match the kernel dimension of $\theta^{\lbrack l\rbrack}$. 
% Finally, the new weights $\hat{\theta}^{\lbrack l\rbrack}$ of layer $l$ are calculated by: $\hat{\theta}^{\lbrack l\rbrack}:=\theta^{\lbrack l\rbrack}\cdot(1+\Delta^{\lbrack l\rbrack})$. 

Let $G(w;\theta)$ be the output image of $G$ parameterized by weights $\theta$. 
For each sketch input $x_{sk}$, we first obtain the generated image $\hat{x}_0=G(w;\theta)$. 
At each refinement step $1\leq t\leq T$, the network $H$ predicts a set of weight offsets $\Delta_t=H(x_{sk},\hat{x}_{t-1})$. 
The new weights $\hat{\theta}_t$ of $G$ at step $t$ are the accumulated modulation across all previous steps \cite{alaluf2022hyperstyle}: 
%% Equation (6)
\begin{equation}
\label{eq:6}
\hat{\theta}_t:=\theta \cdot \left(1+\sum_{i=1}^t\Delta_i\right).
\end{equation}

The new generated image is updated by $\hat{x}_t=G(w;\hat{\theta}_t)$ and is used to predict weight offsets at the next step. 
We apply the same $\mathcal{L}_\mathrm{sim}$ loss in Eq. \eqref{eq:1} to train $H$, and the loss is calculated at each refinement step. 
Again, note that the $S$, $E$, and $G$ networks are frozen during training. 
The total refinement step for each sketch input is set to $T=3$ in both training and testing. 
At test time, the final generated image is obtained at the last step. 
For more details on the architecture and implementation of the refinement network $H$, please refer to the literature \cite{alaluf2022hyperstyle}. 
% Figure 2
\input{figures/tex_files/Fig_2}

%% file: figures/tex_files/Fig_2.tex
%Figure 2
\begin{figure}[!t]
\centering
\includegraphics[width=0.95\linewidth]{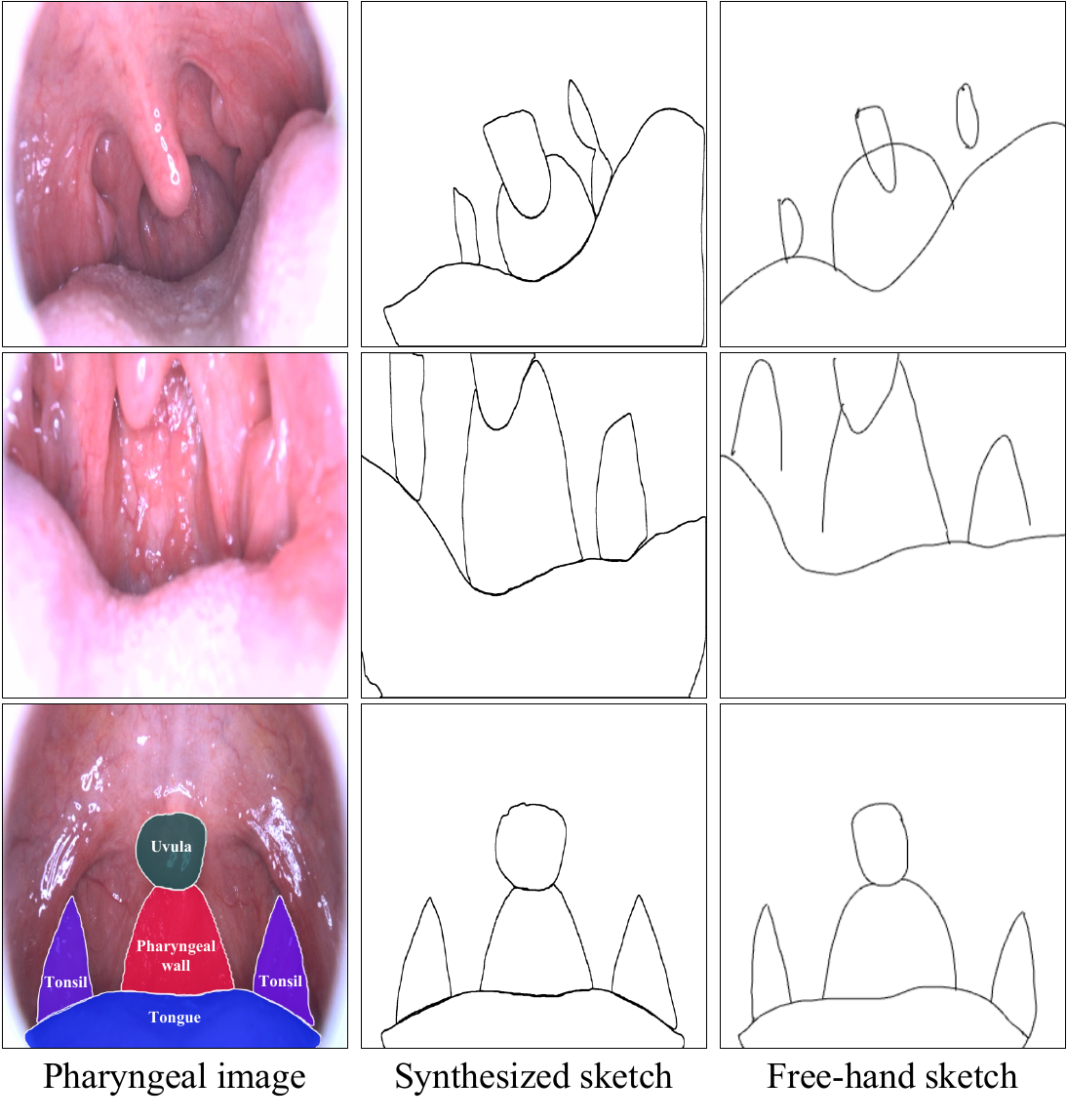}
\caption{
    Samples from the pharyngeal, synthesized sketch, and free-hand sketch datasets. An example of the predicted masks of four pharyngeal objects (i.e., uvula, pharyngeal wall, tonsil, and tongue) is illustrated in the bottom-left image. The sketch images depict these objects.
}
\label{fig:fig_2}
\end{figure}

%% file: 04_Experiment_settings.tex
% Table I
\input{tables/table_I}
\subsection{Datasets}
In this study, we analyzed pharyngeal images and created various datasets. 
Fig. \ref{fig:fig_2} illustrates some samples from these datasets. 
The specific of each dataset is described as follows: 
\subsubsection{Pharyngeal dataset} We collected a large number of high-quality 60,000 pharyngeal RGB images from over 5,000 patients. 
These images were acquired using a camera specifically designed for taking pharyngeal images. 
Eventually, all images were resized to $256 \times 256$ pixels. 

\subsubsection{Segmentation dataset} Another 2,000 pharyngeal images were annotated with segmentation annotations by experienced experts. 
The labels include four classes namely: \textit{uvula}, \textit{pharyngeal wall}, \textit{tonsil}, and \textit{tongue}. 
From this data, 1,600 and 400 images were used for training and validating the sketch synthesizer $S$, respectively. 

\subsubsection{Synthesized sketch dataset} After the training, $S$ is applied to synthesize 60,000 corresponding sketch images from the above pharyngeal dataset. 
Given the predicted segmentation masks from a pharyngeal image with confidence scores $\geq\delta$, a synthesized sketch is formed by extracting the object contours from all instance masks. 
We set $\delta=0.35$ in our work. 

\subsubsection{Free-hand sketch dataset} The free-hand sketches are drawn on blank backgrounds by observing the reference pharyngeal images. 
Here, 12 users (exclusive from the segmentation annotators) were asked to draw the boundaries of four pharyngeal class objects as in the segmentation dataset (i.e., uvula, pharyngeal wall, tonsil, and tongue), resulting in 495 sketch images. 
Note that the reference pharyngeal images used to create this dataset are exclusive from the pharyngeal dataset. 

Typically, free-hand sketches contain different levels of abstractions and drawing styles from multiple users. 
Thus, these sketches are usually misaligned with the objects of the referenced pharyngeal images (see Fig. \ref{fig:fig_2}). 

\subsection{Evaluation methods}
We evaluate the generated results from the free-hand sketch dataset in two aspects: \textit{quality} (e.g., clearness, naturalness of outputs) and \textit{originality} (e.g., structure similarity between outputs and reference pharyngeal images), similar to \cite{cap2023practical}. 
To measure the quality aspect, we used the subjective mean opinion score (MOS) and non-reference image quality assessment (NR-IQA) criteria, including FID \cite{heusel2017gans}, TReS \cite{golestaneh2022no}, and CLIP-IQA \cite{wang2023exploring}. 

For the originality aspect, we measured the mean DICE score between predicted segmentation masks of the reference image $x$ and the generated image $\hat{x}$ by applying the trained sketch synthesizer $S$. 
Along with that, we also used the full-reference image quality assessment (FR-IQA) criteria, including LPIPS \cite{zhang2018unreasonable}, DISTS \cite{ding2020image}, and STLPIPS \cite{ghildyal2022shift}. 
Note that the \textit{originality} aspect in this case is not ideal since free-hand sketches are not precisely aligned with the reference pharyngeal object boundaries, however, we still consider this a practical scenario. 

\subsection{Training details}
\subsubsection{The sketch synthesizer} The synthesizer $S$ is based on the SOLOv2 instance segmentation model \cite{wang2020solov2}. 
We trained $S$ on the training set of the segmentation dataset (i.e., 1,600 images) using the default hyperparameters of SOLOv2 with ResNet101 \cite{he2016deep} backbone. 
After the training, we achieved the mean DICE score of 86.49 on the validation set. 

\subsubsection{The generator} The pharyngeal dataset is used to train our generator $G$. 
The output size of $G$ is set to $256 \times 256$ pixels. 
We applied training with the default configurations from the StyleGAN2 literature \cite{karras2020analyzing} and finished the training after 100 epochs. 

\subsubsection{The encoder and refinement networks} The pharyngeal and synthesized sketch datasets are used to train both $E$ and $H$. 
We trained $E$ together with the latent discriminator $D$ in an adversarial manner using the losses in Eq. \eqref{eq:3} and \eqref{eq:5}, while $H$ is trained with the loss from Eq. \eqref{eq:1}. 
% We set the coefficients $\lambda_{l2}, \lambda_{lpips}, \lambda_{cos}$, $\gamma$, and $\lambda_{adv}$ of those losses to $1.0, 0.8, 0.5, 10, 0.1$, respectively. 
We set the coefficients of those losses to $\lambda_{l2}=1.0, \lambda_{lpips}=0.8, \lambda_{cos}=0.5, \gamma=10$, and $\lambda_{adv}=0.1$. 
Note here again that $H$ is trained after we have done training $E$ and both models were trained for 100 epochs. 
% We used the same hyperparameters from \cite{tov2021designing} and \cite{alaluf2022hyperstyle} for training $E$ and $H$, respectively. 
% Both models were trained for 100 epochs. 

\subsubsection{Comparison models} For comparison purposes, we also trained four other sketch-to-image generation models using pix2pix \cite{isola2017image}, CycleGAN \cite{zhu2017unpaired}, UNIT \cite{liu2017unsupervised}, and U-GAT-IT \cite{kim2020u}. 
We trained these models on the same pharyngeal and synthesized sketch datasets with their default configurations. 
All models were trained for 100 epochs. 

%% file: tables/table_I.tex
\begin{table*}[t]
\centering
\caption{Evaluation results of quality and originality aspects on generated images from the free-hand sketch dataset}
\label{tab:table_1}
\begin{threeparttable}

\begin{tabular}{@{}lllll@{\hspace{5pt}}lllll@{}}
\toprule
\multirow{2}{*}{Method}      & \multicolumn{4}{c}{Quality aspect}                                             &  & \multicolumn{4}{c}{Originality aspect}                                        \\ \cmidrule(lr){2-5} \cmidrule(l){7-10} 
                             & FID (↓)        & MOS* (↑)           & TReS (↑)            & CLIP-IQA (↑)       &  & DICE (↑)       & LPIPS (↓)          & DISTS (↓)          & STLPIPS (↓)        \\ \cmidrule(r){1-5} \cmidrule(l){7-10} 
pix2pix \cite{isola2017image}                      & 105.02         & 2.77±1.10          & 65.28±6.07          & 0.52±0.06          &  & 46.58          & 0.42±0.06          & 0.24±0.03          & 0.36±0.07          \\
CycleGAN \cite{zhu2017unpaired}                     & 87.06          & 1.99±0.96          & 59.78±15.44         & 0.46±0.12          &  & 38.19          & 0.48±0.11          & 0.28±0.05          & 0.41±0.10          \\
UNIT \cite{liu2017unsupervised}                         & 117.80         & 2.18±0.92          & 63.27±7.08          & 0.39±0.09          &  & 48.39          & 0.50±0.09          & 0.27±0.03          & 0.40±0.09          \\
U-GAT-IT \cite{kim2020u}                     & 118.83         & 1.42±0.70          & 50.53±9.38          & 0.36±0.09          &  & 19.03          & 0.58±0.12          & 0.33±0.06          & 0.47±0.10          \\ \cmidrule(r){1-5} \cmidrule(l){7-10} 
Sketch2MedI w/o $H$ (\textbf{proposed}) & \textbf{38.39} & N/A                & 76.91±6.47          & 0.62±0.07          &  & 56.53          & 0.36±0.05          & \textbf{0.20±0.03} & 0.30±0.07          \\
Sketch2MedI (\textbf{proposed})       & 40.83          & \textbf{3.99±0.98} & \textbf{78.04±6.28} & \textbf{0.63±0.07} &  & \textbf{58.28} & \textbf{0.35±0.05} & \textbf{0.20±0.03} & \textbf{0.29±0.07} \\ \bottomrule
\end{tabular}
\begin{tablenotes}[flushleft]
    \item[*] The MOS of the \say{Sketch2MedI w/o $H$} model is unavailable since it is part of the proposal. 
    \item[$\dag$] \textbf{Bold} text indicates the best performance. 
\end{tablenotes}
\end{threeparttable}
\end{table*}

%% file: 05_Result_Discussion.tex
For the subjective MOS evaluation, 20 users participated in scoring the quality (e.g., clearness, naturalness) of the generated images from pix2pix, CycleGAN, UNIT, U-GAT-IT, and our Sketch2MedI models on a scale of 1 to 5 (bad to excellent). 
Because of the time-consuming process, we randomly selected 80 images from the free-hand sketch dataset for the MOS test. 
Subsequently, each user evaluated a total of 400 images, and the final score for each model was averaged across all participants. 
Table \ref{tab:table_1} shows the numerical evaluations of the quality and originality aspects. 
To confirm the effect of the refinement network $H$, we also included the results of our proposal without using $H$, noted as \say{Sketch2MedI w/o $H$} in Table \ref{tab:table_1}. 
Here, for all cases, we refer to \say{Sketch2MedI} as our complete model with both $E$ and $H$ employed. 
Overall, the use of network $H$ is effective in improving the quality and originality aspects. 
Visual comparisons of the generated images from pix2pix, CycleGAN, UNIT, U-GAT-IT, and our Sketch2MedI are shown in Fig. \ref{fig:fig_3}. 
% Figure 3
\input{figures/tex_files/Fig_3}

Due to the large distribution shift between the synthesized and free-hand sketches, generated images by pix2pix, CycleGAN, UNIT, and U-GAT-IT are often deformed and unrealistic. 
From the experimental results, the outputs of the supervised pix2pix model somewhat align with the sketch edges. 
However, its overall quality remains unsatisfactory. 
Concurrently, the unpaired image translation methods (i.e., CycleGAN, UNIT, and U-GAT-IT) severely failed to generalize on the free-hand sketches. 
These methods were not able to learn meaningful representations from sketch data, thus yielding inferior results (Table \ref{tab:table_1}, Fig. \ref{fig:fig_3}). 

On the contrary, thanks to the ability to encode the input sketches into StyleGAN’s latent space which has expressive and semantically rich representations, the Sketch2MedI model is robust against the distribution shift, consistently ensuring the generation of visually pleasing and high-quality results. 
Our method performed efficiently under various user drawing styles and significantly surpassed all other methods in terms of the quality aspect (FID, MOS, and NR-IQA). 
We believe the high-quality medical image generation of the Sketch2MedI method could be a practical application in patient communication, as previously discussed. 
% Here, using $H$ slightly reduced the FID scores. 
% Since FID is calculated based on the distribution between the training pharyngeal data and the generated data from sketches, we believe the slight reduction of FID is due to the distribution shift between synthesized and free-hand sketches. 

For the originality aspect, our method also achieved superior results to other methods (in DICE and FR-IQA). 
The utilization of $H$ has been proven to improve originality preservation compared to when using the encoder $E$ alone (Sketch2MedI w/o $H$ in Table \ref{tab:table_1}). 
In addition, the same performance trend has been observed where $H$ helped to improve the mean DICE score by 5.03 points (from 71.59 to 76.62) on 6,000 test synthesized sketches. 
Although this originality evaluation is considerably not ideal as the free-hand sketches are not well-aligned with target image objects (see Fig. \ref{fig:fig_2}), these results suggest the possibility of utilizing our Sketch2MedI for downstream tasks such as sketch-based medical image retrieval. 

Despite achieving practical results, our Sketch2MedI occasionally generated images with loosened sketch information. 
For instance, in Fig. \ref{fig:fig_3} (last row, red arrows), the results from Sketch2MedI do not precisely express the intended tonsil shapes (i.e., generated smaller shapes) compared to other models. 
This drawback might be because the sketch data is encoded into the abstract latent space before the generation, causing the inadequacy in capturing sketch information. 
We believe introducing some constraints during training can mitigate this issue and intend to improve our model in future works. 

%% file: figures/tex_files/Fig_3.tex
%Figure 3
\begin{figure*}[!t]
\centering
\includegraphics[width=0.95\textwidth]{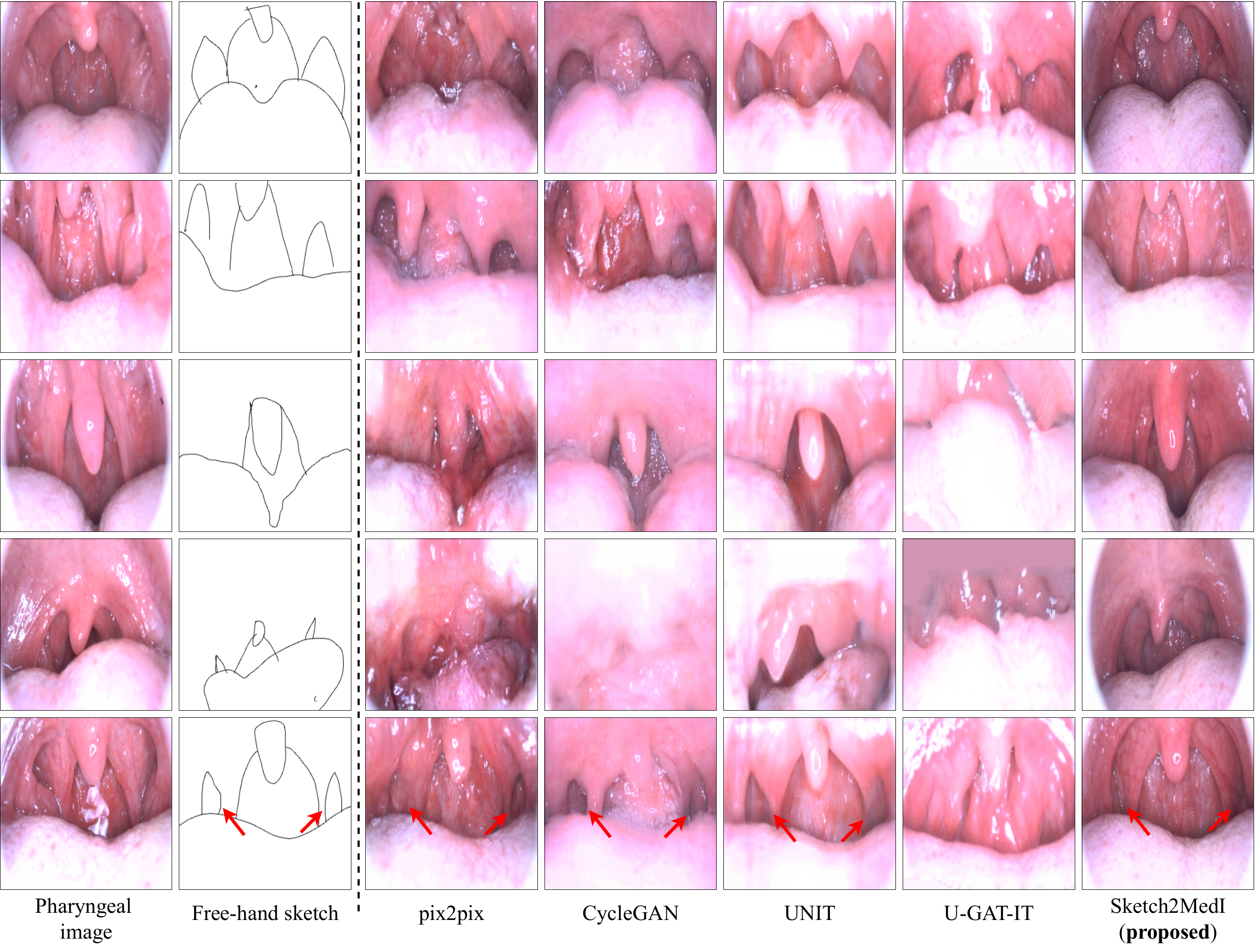}
\caption{
    Visual comparison of the generated images from the free-hand sketch dataset by pix2pix \cite{isola2017image}, CycleGAN \cite{zhu2017unpaired}, UNIT \cite{liu2017unsupervised}, U-GAT-IT \cite{kim2020u}, and our proposed Sketch2MedI models. 
}
\label{fig:fig_3}
\end{figure*}

%% file: 06_Conclusion.tex
In this study, we proposed the practical Sketch2MedI model for realistic free-hand sketch to medical image generation. 
By encoding sketch inputs into StyleGAN’s latent space which has semantically rich representations and then generating medical images from it, the Sketch2MedI proved to be effective in generalizing to the free-hand sketches even though it was trained entirely on the different synthesized sketch dataset. 
Our proposal not only generates high-quality and realistic images but also exhibits its applicability to useful utilization in the field of medical imaging. 